\documentclass[runningheads]{llncs}
\usepackage{graphicx}
\usepackage{amsmath,amssymb} 
\usepackage{color}
\usepackage{multirow}
\usepackage{floatrow}
\newfloatcommand{capbtabbox}{table}[][\FBwidth]
\begin{document}
\pagestyle{headings}
\mainmatter

\def\ACCV22SubNumber{134}  

\title{A Unified Framework with Meta-dropout for Few-shot Learning} 

\author{Shaobo Lin \and
	Xingyu Zeng \and
	Rui Zhao}

\institute{Sensetime Research\\
	\email{\{linshaobo,zengxingyu,zhaorui\}@sensetime.com}\\}

\maketitle


\begin{abstract}
	Conventional training of deep neural networks usually requires a substantial amount of data with expensive human annotations. In this paper, we utilize the idea of meta-learning to  explain two very different streams of few-shot learning, \emph{i.e.}, the episodic meta-learning-based and pre-train finetune-based few-shot learning, and form a unified meta-learning framework. 
	In order to improve the generalization power of our framework, we propose a simple yet effective strategy named meta-dropout, which is applied to the transferable knowledge generalized from base categories to novel categories.  The proposed strategy can effectively prevent neural units from co-adapting excessively in the meta-training stage. Extensive experiments on the few-shot object detection and few-shot image classification datasets, \emph{i.e.}, Pascal VOC, MS COCO, CUB, and mini-ImageNet, validate the effectiveness of our method.
\end{abstract}

\section{Introduction}
\label{submission}

Deep Neural Networks (DNNs) have achieved great progress in many computer vision tasks ~\cite{ren2016faster,dai2016r,redmon2017yolo9000,lin2017feature}. However, the impressive performance of these models largely relies on a large amount of data as well as expensive human annotation. When the annotated data are scarce, DNNs cannot generalize well to testing data especially when the testing data belongs to different classes of the training data. In contrast, humans can learn to recognize or detect a novel object quickly with only a few labeled examples. Due to some object categories naturally have few samples or their annotations are extremely hard to obtain, the generalization ability of conventional neural networks is far from satisfactory.

Few-shot learning, therefore, becomes an important research topic to achieve better generalization ability by learning from only a few examples. 
The mainstream few-shot learning approaches consists of episodic  approaches ~\cite{kang2019few,yan2019meta,wang2019meta,fan2020few} and pretrain-finetune based approaches~\cite{dhillon2019baseline,chen2019closer,wang2020frustratingly}. Episodic meta-learning encapsulates the training samples into an episode~\cite{vinyals2016matching} to mimic the procedure of few-shot testing. Pre-train finetune-based methods 
are composed of the pre-training stage and fine-tuning stage, the former is responsible for obtaining a good initialization point from base classes, and the latter adapts the pre-trained model to a specific task, respectively. In order to transfer knowledge from base data to novel ones, both methods are trained with two stages, where the data-sufficient base classes and the data-scarce novel classes are used separately. However, there is no framework to unify the two very different streams, hindering exploring the common and eccentric problem of few-shot learning.

\begin{figure}[t]
	\begin{center}
		\centerline{\includegraphics[width=\columnwidth]{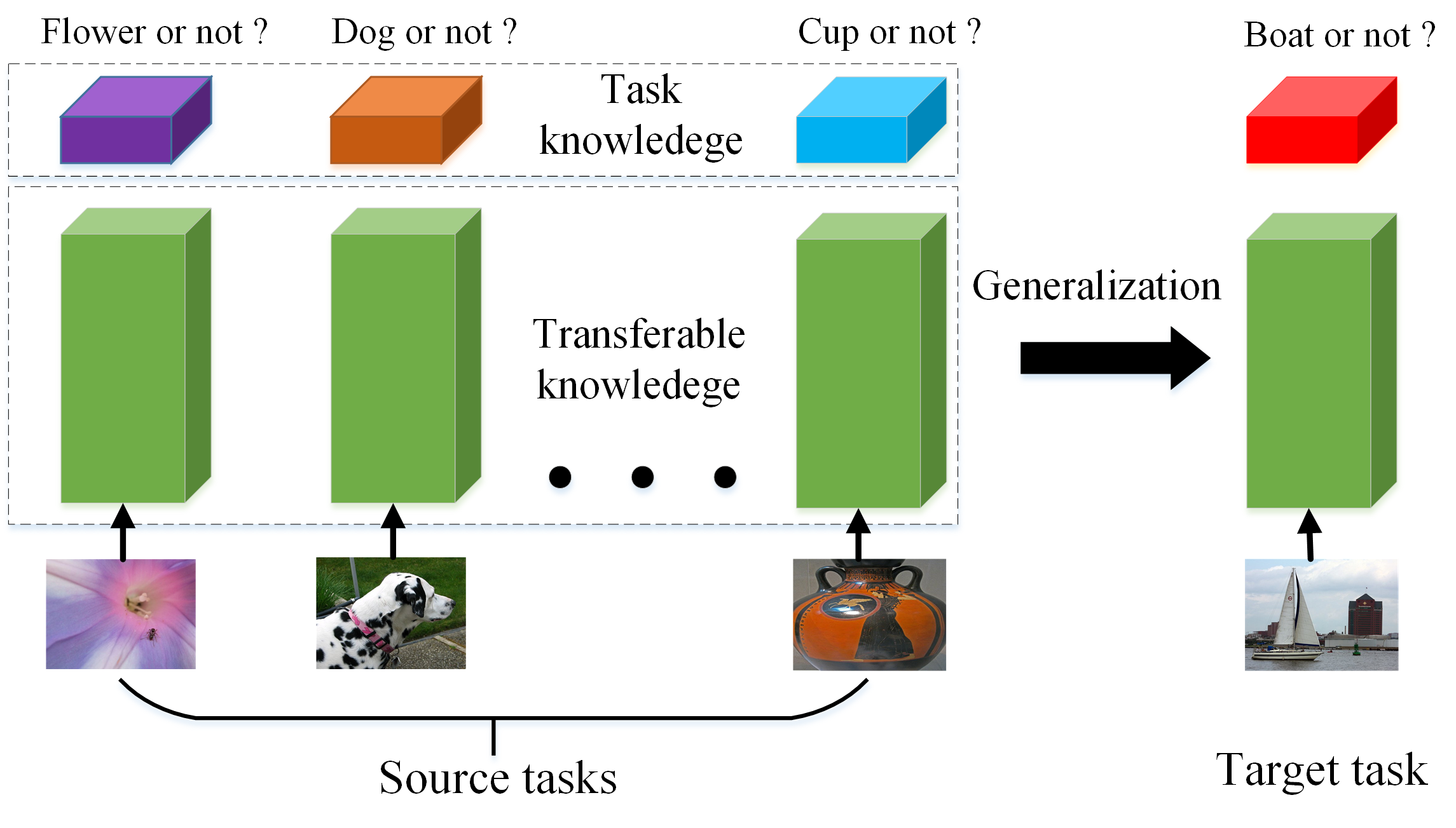}}
		\caption{The generalization power of transferable knowledge across different source tasks is the key for few-shot learning, in which transferable knowledge is adapted to the target task.}
		\label{fig1}
	\end{center}
\end{figure}

In this paper, we incorporate episodic meta-learning-based which is denoted as episode-based for simplification, and pre-train finetune-based few-shot methods into one unified optimization framework based on the idea of meta-learning. The framework consists of a novel reformulated meta-training stage and a meta-testing stage. In the meta-training stage, our framework considers the common elements of few-shot learning, including meta-knowledge, task-knowledge, meta-loss, task-loss, and the distribution of the data and tasks. In the meta-testing stage, the final model for novel tasks is obtained based on the learned model.

As shown in Fig.\ref{fig1}, a deep model can be divided into two components, called task-specific knowledge and transferable knowledge. The former  represents the last fully-connected classifier for specific categories, the latter is the well-learned feature representation which is needed to be generalized to novel tasks. In the different approaches, meta-knowledge can be different types, such as the frozen feature in TFA~\cite{wang2020frustratingly} and the initialization point of backbone in FSCE~\cite{sun2021fsce}, etc. Therefore, it is crucial to improve the generalization power of transferable knowledge if we want to apply it from source tasks to novel tasks in few-shot learning. To achieve this goal, we propose a simple yet effective strategy, named meta-dropout. Because our unified framework can integrate two very different streams of few-shot learning and identify which part of a model is the transferable knowledge, the proposed meta-dropout can be easily applied to existing few-shot models. We select several different methods from the above two streams as the baselines to validate the correctness of our framework and the effectiveness of our meta-dropout. By utilizing meta-dropout, our model demonstrates great superiority towards the current excellent methods on few-shot object detection and few-shot image classification tasks. Dropout is not a new idea. However, we use it to solve a new problem (few-shot learning) and we provide more insights about how to use it, which are the major novelties. Our overall contributions can be summarized as three-fold: 

\begin{itemize}
	
	\item We utilize the idea of meta-learning to  explain two different streams of few-shot learning, \emph{i.e.}, the episodic meta-learning-based and pre-train finetune-based few-shot learning, and form a unified meta-learning framework.
	
	\item We propose a simple yet effective strategy, named meta-dropout, to improve the generalization power of meta-knowledge in our framework. 
	
	\item Experiments of baselines from different streams evaluate the effectiveness of our approach on the few-shot object detection and image classification datasets, \emph{i.e.}, Pascal VOC, MS COCO, CUB, and mini-ImageNet. 
	
\end{itemize}

\section{Related Work}

The episode-based and pre-train finetune-based methods are two existing mainstream methods in few-shot learning.  The differences between the episode-based and pre-train finetune-based methods include the training pipeline (normal training or episodic training) and the distribution of datasets during training (one overall task or multiple serial tasks). 

\subsection{Episode-based Few-shot Learning}
Few-shot learning is an important yet unsolved task in computer vision~\cite{hariharan2017low,koch2015siamese,vinyals2016matching,tokmakov2019learning}. Nowadays, the meta-learning strategy which is called “learning-to-learn” has become an increasingly popular solution. The goal of meta-learning is to obtain task-level meta-knowledge that helps the model quickly generalize across all tasks~\cite{andrychowicz2016learning,munkhdalai2017meta,santoro2016one,thrun1998lifelong}. Recent methods for few-shot learning usually extract meta-knowledge from a set of auxiliary tasks via the episode-based strategy~\cite{vinyals2016matching}, where each episode contains $C$ classes and $K$ samples of each class, \emph{i.e.}, $C$-way $K$-shot. 


In few-shot image classification,~\cite{vinyals2016matching} proposed Matching Networks to find the most similar class for the target image among a small set of labeled samples. Prototypical Networks (PN)~\cite{snell2017prototypical} extended Matching Networks by producing a linear classifier instead of the weighted nearest neighbor for each class. 
The cosine similarity-based classifier further enhanced the discriminative power of the trained model~\cite{chen2019closer,gidaris2018dynamic}. Relation Network (RN)~\cite{sung2018learning} used a neural network to learn a distance metric, in which the unlabeled images could be classified according to the relation scores between the target sample and a few labeled images. 
Graph Neural Network (GNN)~\cite{kim2019edge,gidaris2019generating} was also utilized to model relationships between different categories. 
In few-shot object detection,~\cite{kang2019few} applied a feature re-weighting module to a single-stage object detector (YOLOv2) with the support masks as inputs.~\cite{yan2019meta} introduced a Predictor-head Remodeling Network (PRN) that shared its backbone with Faster/Mask R-CNN. To disentangle the category-agnostic and category-specific components in a CNN model, ~\cite{wang2019meta} proposed a weight prediction meta-model for predicting the parameters of category-specific components from few samples.
~\cite{karlinsky2019repmet} proposed a new module for distance metric learning (DML) that could be used as the classification head combined with the standard object detection model.~\cite{fan2020few} introduced a few-shot object detection method, which consisted of an attention-rpn, a multi-relation detector, and a contrastive training strategy. 

\subsection{Pre-train Finetune-based Few-shot Learning}
Pre-train finetune-based approaches are basic yet ignored in few-shot learning due to the excellent performance of episode-based methods. However, some simple pre-train finetune-based methods turn out to be more favorable than many episode-based works~\cite{chen2019closer,dhillon2019baseline,chen2021meta}.~\cite{chen2019closer} introduced a pre-train finetune baseline with a distance-based classifier, achieving a competitive performance with state-of-the-art episode-based classification approaches. ~\cite{chen2021meta} explored a simple process: meta-learning over a whole classification pre-trained model. This simple method achieves competitive performance to state-of-the-art methods on standard benchmarks. 
By using the proposed regularization, the standard detectors, such as SSD~\cite{liu2016ssd} and FRCNN~\cite{ren2016faster}, were fine-tuned for few-shot problems. 
Furthermore,~\cite{wang2020frustratingly} demonstrated that only fine-tuning the last layer of existing models was crucial to the few-shot object detection task. Such a simple approach outperforms the episode-based approaches by 2 to 20 points and even doubles the accuracy of the prior works on current benchmarks. 

\begin{figure}[t]
	\begin{center}
		\centerline{\includegraphics[width=\columnwidth]{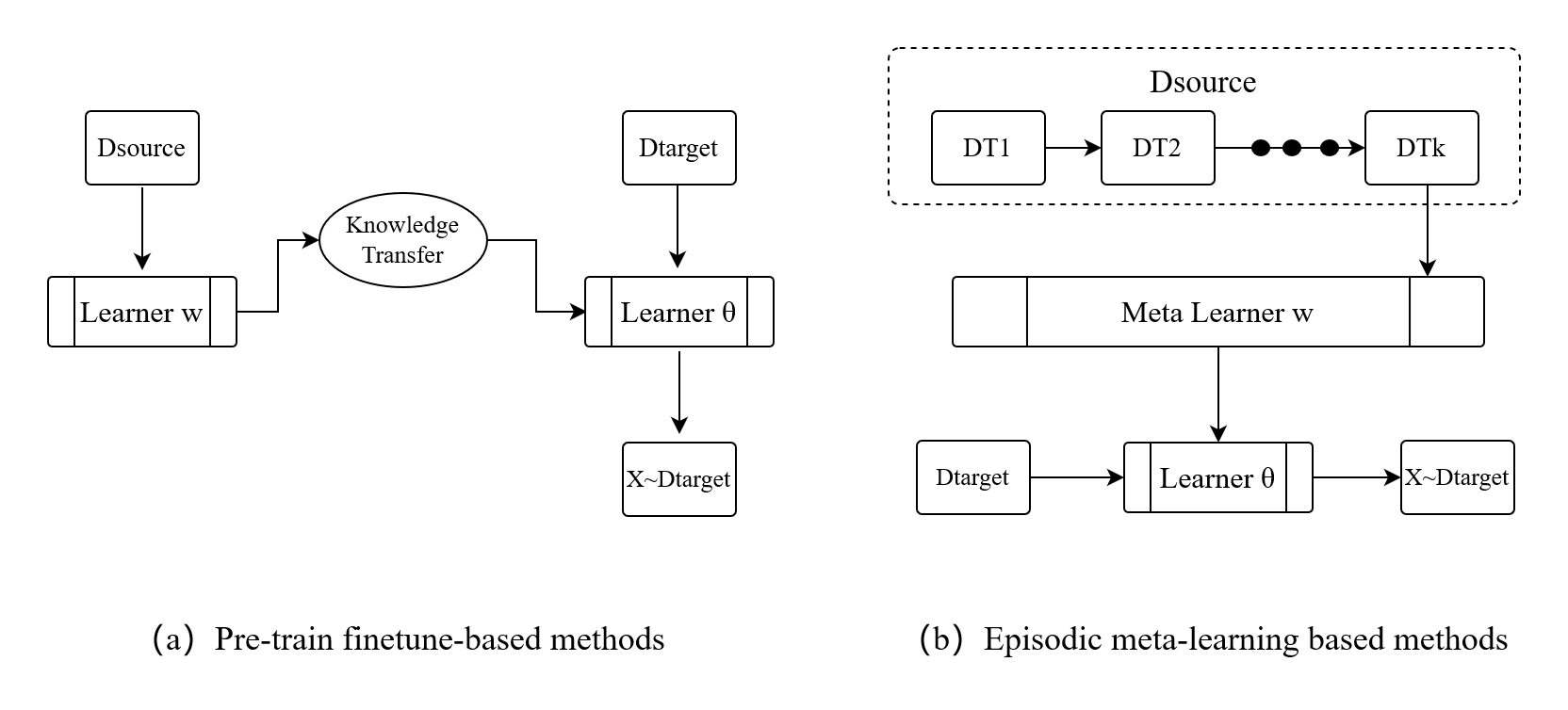}}
		\caption{The optimization processes of pre-train finetune-based and episodic meta-learning-based methods.}
		\label{fig2}
	\end{center}
\end{figure}

\section{Method}

\subsection{The Solution of Few-shot problems}
Common supervised learning problems based on abundant training data can be solved by minizing the Equ (1), in which $\Theta$ is the trainable parameters of neural networks, and x is input data sampled from $p(D)$. While few-shot tasks are limited in the number of samples, optimizing it directly is likely to results in the overfitting of $\Theta$ due to its high-dimensional property.

\begin{equation}
	\underset{\Theta}{min}\underset{x \sim p(D)}{E}{L(\Theta; x)}
\end{equation}   

Few-shot learning aims to solve learning problems with just a few training examples. To achieve this goal, the common solution is to reduce the learnable dimension of $\Theta$, and thus the Equ (1) can be re-written as Equ (2), in which $\Theta$ = [$\theta$,$w$]. 	 
$w$ represents the useful foundation for few-shot learning, such as a good initailization point or a well-learned feature representation, which is obtained from source tasks. $\theta$ are updated on target tasks based on the learned $w$.

\begin{equation}
	\underset{\theta}{min}\underset{x \sim p(D)}{E}{L(\theta;x | w)}
\end{equation}


Given a labeled source dataset $D_{source}$, there are $C_{source}$ source classes with a large number of images in each class. Novel dataset $D_{target}$ with novel classes $C_{target}$ consists of a few samples in each class. $C_{source}$ and $C_{target}$ do not have overlapping categories. The learning goal is adapting the model from $D_{source}$ to $D_{target}$. $D_{source}$ and $D_{target}$ are used to optimize $w$ and $\theta$ respectively. The $C$-way $K$-shot few-shot setting that is used for evaluating the performance of a few-shot model means $D_{target}$ has $C$ novel categories and each novel category has $K$ images.

Existing few-shot learning methods consist of two different streams of categories, \emph{i.e.}, the episodic meta-learning-based and pre-train finetune-based methods. However, both streams can be explained by the above formulation as shown in Fig.\ref{fig2}. $w$ in pre-train finetune-based methods, like TFA~\cite{wang2020frustratingly}, is its backbone which is transferred to the second stage for initialization. $\theta$ in TFA means the last fully-connected layer that is optimized for novel tasks based on the frozen $w$. $w$ in Meta R-CNN~\cite{yan2019meta}, which is a representative episodic meta-learning-based method, provides a backbone with good generalization ability.  $\theta$ in Meta R-CNN is the trainable backbone and the last fully-connected regression and classification layers for novel tasks. 

\subsection{The Unified Meta-learning Framework}
In order to explain the learning process of existing few-shot methods, we propose a unified meta-learning framework to re-formulate Equ (2) as Equ (3), in which the learning goal is to obtain a general transferable knowledge via optimizing the expectation of many source tasks and source datasets. 
Suppose $p(T)$ presents the distribution of source tasks, in which $T$ consists of infinite basic tasks, and $p(D_t)$ is the distribution of the training dataset $D_t$ of task $t$, in which the number of the training data is unlimited. 
$w$ is used to specify meta-knowledge, which is the transferable parameters among different tasks. 
$\theta$ is task-specific knowledge, called task-knowledge. $L$ is the loss function. The overall optimization goal is:

\begin{equation}
	\underset{w}{min}\underset{t \sim p(T)}{E}\underset{x \sim p(D_t)}{E}{L(w;\theta_t;x)}
\end{equation}

The goal is to let $w$ get good performance for each task $T$=\{$D$,$L$\}. 
Our meta-learning framework consists of a meta-training phase and a meta-testing phase.

\subsubsection{Meta-training}
The goal of meta-training is formulated below: 

\begin{equation}
	w*=\arg \underset{w}{max}\log p(w|T)
\end{equation}

In our meta-training stage, the formulas of optimization are:

\begin{equation}
	w* = \underset{w}{\arg min}\underset{t \sim p(T)}E\underset{x \sim p(D_t)}EL^{meta}(\theta^{*(i)}(w),w)
\end{equation}

\begin{equation}
	s.t. \theta^{*(i)}(w)= \underset{\theta}{\arg min}\underset{D_{ij} \sim p(D_{i})}EL^{task}(\theta^i,w)
\end{equation}
$w$ is meta-knowledge and $w$* is the learned $w$ during the iterations of tasks. $\theta$ represents task-knowledge and $\theta$* is the learned task-knowledge that is used to optimize $w$. $L^{task}$ is used to optimize the task-specific information and $L^{meta}$ is for generating the best meta-knowledge $w$*. $L^{meta}$ called meta-loss represents which $w$ is good, and $L^{task}$ represents which model is good for a specific task, called task-loss. 
From formulas (5) and (6), we can obtain the episode-based and pre-train finetune-based frameworks in the following sections.

\subsubsection{Meta-testing}
Pick a set of $Q$ target tasks to apply the learned $w$ to obtain a task-specific model. 
\begin{equation}
	D_{target}=\{(D_{target}^{train},D_{target}^{test})^{(i)}\}_{i=1}^{Q}
\end{equation}

\begin{equation}
	\theta^{*(i)}=\arg \underset{\theta}{max}\log p(\theta|w^{*},D_{target}^{train(i)})
\end{equation}
$D_{target}^{train}$ and $D_{target}^{test}$ are built from the novel dataset $D_{target}$ with a small scale of classes and samples. 
$D_{target}^{train}$ corresponds to all annotated data and is used to further optimize the model for the specific tasks. $D_{target}^{test}$ is the real novel test data.
$w$* is the learned best meta-knowledge that is used for obtaining $\theta$*. $\theta$* is the best task-knowledge for the final prediction.
\subsubsection{Episode-based Meta-learning Framework}

Episodic meta-training based methods is optimized in the form of episode, and each episode is corresponding to a specific task. Each episode consists of two parts: support set and query set, which are presented by $D_{source}^{train}$ and $D_{source}^{val}$ in the following formula. $D_{source}^{train}$ and $D_{source}^{val}$ are built from the $D_{source}$.

\begin{equation}
	D_{source}=\{(D_{source}^{train},D_{source}^{val})^{(i)}\}_{i=1}^{M}
\end{equation}

Consistent with our unified meta-learning framework, by solving the above equations using the episodic dataset  $D_{source}^{train}$ and $D_{source}^{val}$, the formulas of optimization are:

\begin{equation}
	w*= \underset{w}{\arg min}\sum_{i = 1}^{M}\sum_{j = 1}^{N}L^{meta}(\theta^{*(i)}(w),w,D_{source}^{val(ij)})
\end{equation}

\begin{equation}
	s.t. \theta^{*(i)}(w)= \underset{\theta}{\arg min}\sum_{j = 1}^{N}L^{task}(\theta^i,w,D_{source}^{train(ij)})
\end{equation}
$M$ is the number of source tasks sampled from $p(T)$ and $N$ means the number of samples in each task. When building an episode, some categories of data will be randomly selected to build the $D_{source}^{train}$, and some samples with the same classes as the $D_{source}^{train}$ will be selected from the remaining data to build the $D_{source}^{val}$. Specifically, $w$ represents the task-independent network parameters in meta-training stage, and $\theta$ is the task-specific network parameters in meta-testing stage. 
$L^{meta}$ is calculated on the $D_{source}^{val}$ and $L^{task}$ is computed based on $D_{source}^{train}$. 

There are two changes from our unified framework to the episode-based formula. First, the number of tasks and the number of samples are limited in each episode, which are presented by $M$ and $N$ respectively. Therefore, using more tasks and more samples by increasing $M$ and $N$ in meta-training is important for better performance. Second, $D_{source}^{train}$ and $D_{source}^{val}$ are built from a small set of data from the whole dataset, whose distribution is very different from that of the overall dataset. In order to get excellent results, the gap should be reduced. During episode-based training, we can sample data as diverse as possible, to use $[p(x_1),p(x_2),....,p(x_M)]$ to simulate the real distribution of $p(D_T)$.

\subsubsection{Pre-train Finetune-based Meta-learning Framework}

Pre-train finetune methods have a two-stage training framework which consists of a base-training stage and a fine-tuning stage. In our meta-learning framework, the base training stage is presented by the meta-training stage and the fine-tuning stage is the meta-testing stage.

In the meta-training stage, a set of $M$ source tasks from $p(T)$ and a set of $N$ samples of each task are sampled to learn $w$, thereby the formulas of optimization in pre-train finetune-based methods are:

\begin{equation}
	w*= \underset{w}{\arg min}\sum_{i = 1}^{M}\sum_{j = 1}^{N}L^{meta}(\theta^{*(i)}(w),w,D_{source(ij)})
\end{equation}

\begin{equation}
	s.t. \theta^{*(i)}(w)= \underset{\theta}{\arg min}\sum_{j = 1}^{N}L^{task}(\theta^i,w,D_{source(ij)})
\end{equation}

Consistent with our unified meta-learning framework, a large pre-training dataset $D_{source}$ is used to provide a good initialization for meta-testing. 
$w$ represents the backbone parameters in the meta-training stage. $\theta$ corresponds to the fine-tuning part of the model in the meta-testing phase, which is task-related. The common fine-tuning parts are the final classification and regression layers. 
The gap between the unified framework and the pre-train finetune-based framework is the scale of the training dataset. Increasing $M$ and $N$ plays a very important role in pre-train finetune-based methods.

We explain the differences between our framework and two streams of methods using Table.~\ref{table:111} for easy understanding. Episode-based methods are trained with multiple datasets by using episodic-training. Fine-tune-based methods use one dataset via normal training. Moreover, we formulate the fine-tune-based methods in the meta-learning way. Our unified framework can support two dataset types and two training strategies. There are no notable differences between our uniffed framework and the two streams of methods, but summarizing the key components of these
methods into one framework.
\begin{table*}[ht]
	\setlength{\tabcolsep}{4pt}	
	\footnotesize	
	\begin{center}
		\caption{The differences between our unified framework and the two streams of few-shot methods.}
		\label{table:111}
		\begin{tabular}{cccc}
			\hline\noalign{\smallskip}
			Method& data & training & formulation \\
			\noalign{\smallskip}
			\hline
			\noalign{\smallskip}
			episodic-learning based &multiple datasets&episodic &meta-learning\\
			finetune-based &one dataset&normal&None\\	
			ours&both&both&meta-learning\\
			\hline
		\end{tabular}
	\end{center}
\end{table*}


\subsection{Meta-dropout} 
Because our unified framework can integrate two different streams of few-shot learning and identify which part of a model is the meta-knowledge, the proposed meta-dropout can be easily applied to many different few-shot models. 
Meta-dropout means using the idea of dropout~\cite{srivastava2014dropout} on the meta-training stage which is trained with abundant source datasets for improving the generalization power of meta-knowledge. By applying meta-dropout, we can obtain a new formulation of meta-learning below.

\begin{equation}
	\underset{w}{min}\underset{t \sim p(T)}{E}\underset{x \sim p(D_t)}{E}{L(\theta_t; O(w); x)}
\end{equation}
$O$ is applying meta-dropout on $w$.

The optimization objective is below:

\begin{equation}
	w*=\arg \underset{w}{max}\log p(O(w)|T)
\end{equation}

In the meta-training, the formulas of optimization are:

\begin{equation}
	w* = \underset{w}{\arg min}\underset{t \sim p(T)}E\underset{x \sim p(D_t)}EL^{meta}(\theta^{*(i)}(w),O(w))
\end{equation}

\begin{equation}
	s.t. \theta^{*(i)}(w)= \underset{\theta}{\arg min}\underset{D_{ij} \sim p(D_{i})}EL^{task}(\theta^i,O(w))
\end{equation}

Dropout is not a novel idea for generalizing the model. However, the key question is how to use it to solve what we face to and why we can obtain the better performance. Meta-dropout is used to solve new problems (few-shot object
detection and few-shot classification) and we provide more insights in the following sections about how to use it, which are the major
novelties. Meta-dropout is applied on meta-knowledge $w$, while normal dropout is commonly applied on task-specific knowledge. In order to optimize few-shot models better, meta-knowledge should be more generalized. By using meta-dropout on $w$, the degree of over-fitting to the base classes can be relieved, thereby the model is easier to be adapted to novel classes. That is why our meta-dropout is better than normal dropout in few-shot models. 

It is worth noting that our framework, which presents the important and common mathematical principles of existing few-shot models, is different from the methods that combine pre-training and episodic training in methodology to get a fused detector or classifier~\cite{triantafillou2019meta,chen2021meta}. 
The proposed framework is useful for guiding the improvement of existing few-shot models, such as our meta-dropout on meta-knowledge can be generalized to different types of few-shot models and tasks. Therefore, we do extensive experiments on detection and classification tasks with different kinds of baselines in the experimental section.
Moreover, we can get some insights from this framework. Specifically, other strategies, which are applied to the components of one category, can also be used to that of another. For example, regularization and self-supervised learning can be applied to the meta-knowledge of above two training pipelines for better generalization power. 

The novelty of our framework includes:
\begin{itemize}
	\item Based on the original meta-learning framework, we consider the distribution of source tasks ($T$) and the distribution of data in each task ($D_t$) that can present the common characteristics of the above two methods and explain two different training pipelines in theory. 
	\item Our framework can identify which part of a model is the transferable knowledge, so the meta-dropout can be easily applied to existing few-shot models. 
\end{itemize}

\section{Few-shot Object Detection}
\subsection{Datasets}
We evaluate our methods on Pascal VOC ~\cite{everingham2015pascal,everingham2010pascal} and MS COCO~\cite{lin2014microsoft}. 
In PASCAL VOC, we adopt the common strategy~\cite{ren2016faster,dai2016r,redmon2017yolo9000} that using VOC 2007 test set for evaluating while VOC 2007 and 2012 train/val sets are used for training. Following~\cite{yan2019meta}, 5 out of its 20 object categories are selected as the novel classes, while keeping the remaining 15 ones as the base classes. We evaluate with three different novel/base splits from ~\cite{yan2019meta}, named as split 1, split 2, and split 3. 
Following~\cite{yan2019meta,wang2020frustratingly,sun2021fsce},  we use the mean average precision (mAP) at 0.5 IoU threshold as the evaluation metric, and report the results on the official test set of VOC 2007.
When using MS COCO, 20 out of 80 categories are reserved as novel classes, the rest 60 categories are base classes. The  detection performance with COCO-style AP, AP50, and AP75 for K = 10 and 30 shots of novel categories are reported.

\setlength{\tabcolsep}{0.3pt}
\begin{table}[t]
	
	\begin{center}
		\caption{Comparison with state-of-the-art few-shot object detection methods on VOC2007 test set for novel classes of the three splits. $^\Delta$ represents running on one selected seed, and others are averaged over 10 seeds.  {\color{red}{\bf Red}} and {\bf black} indicate state-of-the-art (SOTA) in the setting of single or multiple run. * means applying our meta-dropout.}
		\label{table:1}
		\begin{tabular}{llccccccccc}
			\hline\noalign{\smallskip}
			&&\multicolumn{3}{c}{split 1} & 	\multicolumn{3}{c}{split 2}&\multicolumn{3}{c}{split 3}\\
			\hline\noalign{\smallskip}
			Model Type&Method/Shot& 1 & 3 & 10 & 1 &3&10&1&3&10\\
			\noalign{\smallskip}
			\hline
			\noalign{\smallskip}
			\multirow{5}*{Episode-based}&FR~\cite{kang2019few}&14.8&26.7&47.2&15.7&22.7&40.5&21.3&28.4&45.9\\
			&MetaDet~\cite{wang2019meta}&18.9&30.2&49.6&{\bf21.8}&27.8&43&{\bf20.6}&29.4&44.1\\
			&Meta R-CNN~\cite{yan2019meta}&19.9&35&51.5&10.4&{\bf29.6}&{\bf45.4}&14.3&27.5&{\bf48.1}\\
			&Meta R-CNN (Our Impl.)&14.7&32.8&{\bf51.9}&13.5&24.3&39.2&15.4&33.8&43.1\\
			&Meta R-CNN* &{\bf24.7}&{\bf37.3}&51.8&16.8&28.9&40.1&20.2&{\bf35.4}&45.7\\
			&CME$^\Delta$~\cite{li2021beyond}&41.5&50.4&60.9&27.2&{\color{red}{\bf41.4}}&46.8&34.3&45.1&51.5\\
			\hline
			\hline
			\noalign{\smallskip}
			\multirow{5}*{Pre-train finetune}&FRCN+ft~\cite{kang2019few}&11.9&29&36.9&5.9&23.4&28.8&5.0&18.1&43.4\\
			&FRCN+ft-full~\cite{yan2019meta}&13.8&32.8&45.6&7.9&26.2&39.1&9.8&19.1&45.1\\
			&TFA~\cite{wang2020frustratingly}&25.3&42.1&52.8&{\bf18.3}&{\bf30.9}&39.5&17.9&34.3&45.6\\
			&TFA (Our Impl.)&22.4&40.3&53.1&15.6&26.7&37.4&16.9&32.3&47.7\\
			&TFA*&{\bf26.3}&{\bf45.6}&{\bf55.8}&15.9&29.9&{\bf40.8}&{\bf20.1}&{\bf36.7}&{\bf49}\\
			&MPSR$^\Delta$~\cite{wu2020multi}&41.7&{\color{red}{\bf51.4}}&61.8&24.4&39.2&47.8&35.6&42.3&49.7\\
			&FSCN$^\Delta$~\cite{li2021few}&40.7&46.5&{\color{red}{\bf62.4}}&{\color{red}{\bf27.3}}&40.8&46.3&31.2&43.7&55.6\\	
			&Retentive R-CNN$^\Delta$~\cite{fan2021generalized}&42.4&45.9&56.1&21.7&35.2&40.3&30.2&43&50.1\\
			&HallucFsDet$^\Delta$~\cite{zhang2021hallucination}&{\color{red}{\bf47}}&46.5&54.7&26.3&37.4&41.2&40.4&43.3&49.6\\
			&FSCE$^\Delta$~\cite{sun2021fsce} (Our Impl.)&40.3&47.8&62.2&18.9&39.6&{\color{red}{\bf49.6}}&35.2&44.8&56.1\\				
			&FSCE$^\Delta$*&44.6&50.3&61.7&24.6&40.8&49.4&{\color{red}{\bf41.1}}&{\color{red}{\bf46.2}}&{\color{red}{\bf57.1}}\\
			
			\hline	
		\end{tabular}
	\end{center}
\end{table}

\setlength{\tabcolsep}{8pt}	
\begin{table*}[t]
	\footnotesize		
	\begin{center}
		\caption{Few-shot object detection performance on MS COCO. {\color{red}{{\bf Red}}} / {\bf black} indicate the best / the second best.* means applying our meta-dropout.}
		\label{table:21}
		\begin{tabular}{ccccccc}
			\hline\noalign{\smallskip}
			&\multicolumn{2}{c}{novel AP}&\multicolumn{2}{c}{novel AP50}&	\multicolumn{2}{c}{novel AP75}\\
			\hline\noalign{\smallskip}
			Method/Shot& 10 & 30 & 10 & 30 & 10 & 30 \\
			\noalign{\smallskip}
			\hline
			\noalign{\smallskip}
			FR~\cite{kang2019few}&5.6&9.1&12.3&19&4.6&7.6\\
			Meta R-CNN~\cite{yan2019meta} &8.7&{\bf12.4}&{\color{red}{\bf19.1}}&{\color{red}{\bf25.3}}&6.6&10.8\\	
			TFA (Our Impl.)&{\bf8.9}&12&16.2&21.2&{\bf8.9}&{\bf12.3}\\
			TFA*&{\color{red}{\bf9.7}}&{\color{red}{\bf12.8}}&{\bf17.5}&{\bf22.3}&{\color{red}{\bf9.8}}&{\color{red}{\bf13.1}}\\
			\hline				
		\end{tabular}
	\end{center}
\end{table*}

\subsection{Implementation Details}
Meta R-CNN which is a representative work of episode-based method is adopted as one of our baselines. We use ResNet-101 as the backbone with the structure of Faster R-CNN~\cite{ren2016faster}.
A simple yet effective pre-train finetune-based method~\cite{wang2020frustratingly}, named TFA, is also our baseline. Faster R-CNN is used as the detector and ResNet-101 with a Feature Pyramid Network~\cite{lin2017feature} is the
backbone. In addition, we use a recent excellent method FSCE~\cite{sun2021fsce} as the new baseline to help our model achieve the state-of-the-art performance. 
During experiments, we find using dropblock~\cite{ghiasi2018dropblock} to implement meta-dropout is better than using normal dropout~\cite{srivastava2014dropout} or spatial dropout~\cite{tompson2015efficient}. 
The training strategies of our methods are the same as the selected baseline.

\subsection{Comparison with State-of-the-art Methods}

Based on Meta R-CNN, we apply meta-dropout to get Meta R-CNN* with batch size 1. We also apply meta-dropout on TFA to get TFA*. Based on the official code, we re-implement Meta R-CNN, TFA and FSCE as our baselines denoted by Our Impl in Table.~\ref{table:1}. Specifically, in episode-based methods, except the novel mAP of the 10-shot setting is comparable to the baseline, our Meta R-CNN* achieves more obvious improvement in all other settings. Notably, Meta R-CNN* can gain 10$\%$ improvement in the setting of split 1 with 1-shot. In pre-train finetune-based methods, our TFA* is able to obtain higher accuracy than the baseline in all settings.
In general, our models get the highest improvement in the setting of 1-shot, followed by 3-shot, and the models get the least improvement in the 10-shot setting. With the decrease of the number of novel samples, meta-knowledge in meta-training which presents the power of generalization becomes more important. Therefore, using meta-dropout to improve the generalization power of meta-knowledge can help the models achieve higher improvement in the setting of fewer shots.  Due to the few numbers of novel samples in the second training stage, we use multiple random seeds to get more stable results, as shown in Table.~\ref{table:1}. While comparing to recent methods on Pascal VOC, we only use a certain seed to obtain the results, in order to be consistent with other methods. By applying our meta-dropout to FSCE, our method FSCE* can significantly outperform the baseline and achieve the state-of-the-art performance. In addition, our method does not hurt the performance of base classes which is shown in Table.~\ref{table:100} in appendix.

Few-shot detection results of 10-shot and 30-shot setups for MS COCO are shown in Table.~\ref{table:21}.	
Our methods TFA* can achieve about 1$\%$ gain in most metrics.
It shows that the improvement of TFA is lower than
that on PASCAL VOC, since MS COCO is a more challenging dataset.

\subsection{Ablation Study}
\subsubsection{Meta-dropout}
We follow most of the few-shot methods~\cite{yan2019meta,sun2021fsce,li2021beyond} to do ablation studies with the dataset of VOC 1 split. 
We apply meta-dropout on the meta-training stage and get the model as the initialization for meta-testing. Meta-dropout is implemented by dropblock. Meta-dropout on Meta R-CNN can achieve 1$\%$ improvement on average presented in Table.~\ref{table:2}. Meanwhile, meta-dropout can help TFA achieve higher performance (+4$\%$ on average) in all settings in Table.~\ref{table:3}. In Table.~\ref{table:1}, meta-dropout on FSCE brings 2$\%$-3$\%$ improvement on average. All comparison results show that our meta-dropout can improve the baselines in a fair setting.

\begin{table}[t]
	\setlength{\tabcolsep}{4pt}		
	\begin{center}
	\caption{Results of novel mAP on the split 1 of VOC 2007 test set by using meta-dropout on Meta R-CNN with batch size 1 and 4. }
		\label{table:2}
		\begin{tabular}{ccccccc}
			\hline\noalign{\smallskip}
			Setting / Shot (bs=1/4)& 1 & 3 & 10\\
			\noalign{\smallskip}
			\hline
			\noalign{\smallskip}
			Meta R-CNN&23.3/14.7&{\bf37.8}/32.8&49.3/{\bf51.9}\\
			+ meta-dropout&{\bf24.7}/{\bf16.6}&37.3/{\bf32.9}&{\bf51.8}/51.4\\
			\hline
		\end{tabular}
	\end{center}
\end{table}

\begin{table}[t]
	\setlength{\tabcolsep}{4pt}		
	\begin{center}
		\caption{Results of novel mAP on the split 1 of VOC 2007 test set by adopting meta-dropout on TFA. }
		\label{table:3}
		\begin{tabular}{cccc}
			\hline\noalign{\smallskip}
			Setting / Shot& 1 & 3 & 10 \\
			\noalign{\smallskip}
			\hline
			\noalign{\smallskip}
			TFA&22.4&40.3&53.1\\
			+ meta-dropout&{\bf26.3}&{\bf45.6}&{\bf55.8}\\
			\hline
		\end{tabular}
	\end{center}
\end{table}

\begin{table*}[t]
	\setlength{\tabcolsep}{4pt}		
	\begin{center}
		\caption{Results of novel mAP on the split 1 of VOC 2007 test set by adopting meta-dropout (M) and dropout (D).}
		\label{table:4}
		\begin{tabular}{ccccccc}
			
			\hline\noalign{\smallskip}
			&\multicolumn{3}{c}{Meta R-CNN}&\multicolumn{3}{c}{TFA}\\
			\hline\noalign{\smallskip}
			Setting / Shot & 1 & 3 & 10&1&3&10\\
			\noalign{\smallskip}
			\hline
			\noalign{\smallskip}
			Baseline&14.7&32.8&{\bf51.9}&22.4&40.3&53.1\\
			+ M &{\bf16.2}&{\bf32.9}&51.4&{\bf25.1}&{\bf41.9}&53\\
			+ D &13.3&32.5&50&23.1&41.8&{\bf54.8}\\
			+ M$\&$D&14.5&32.5&48.8&25&41&50.5\\
			\hline
		\end{tabular}
	\end{center}
\end{table*}

\begin{table}[b]
	\setlength{\tabcolsep}{3pt}	
	\begin{center}
		\caption{Results of base and novel mAP on the split 1\&2 of VOC 2007 test set by adopting meta-dropout on TFA. }
		\label{table:100}
		\begin{tabular}{ccccccc}
			\hline\noalign{\smallskip}
			&\multicolumn{3}{c}{split 1} & 	\multicolumn{3}{c}{split2}\\
			\hline\noalign{\smallskip}
			Base / Novel & 1 & 3 &10 & 1 & 3&10 \\
			\hline
			\noalign{\smallskip}
			TFA&79.8/37.2&79.1/{\bf44.4}&{\bf79.4}/53.4&79.8/{\bf21.6}&78.6/34.8&78.3/38.4\\
			TFA*&{\bf80}/{\bf39.7}&{\bf79.7}/43.4&79.2/ {\bf55.2}&{\bf80.6}/20.3&{\bf78.8}/{\bf36.6}&{\bf78.7}/{\bf41.8}\\					
			\hline
		\end{tabular}
	\end{center}
\end{table}

\subsubsection{Meta-dropout vs Dropout}
The implementation of dropout and our meta-dropout can be the same. However, meta-dropout is applied on the meta-knowledge in the meta-training stage, while dropout is applied on the meta-testing stage that is similar to the common one-stage training methods. Meta-dropout is for generalization power, while normal dropout is for specific tasks. 
In order to compare meta-dropout with dropout, we adopt meta-dropout, dropout, and meta-dropout$\&$dropout on Meta R-CNN and TFA. 
From the results in Table.~\ref{table:4}, only applying meta-dropout obtains higher accuracy than using dropout or both strategies in most settings. However, only using dropout on TFA can achieve the highest mAP in the 10-shot setting. The reason may be that dropout can help the pre-trained model get a better ability of fitting novel classes when fine-tuning with relatively more data. 
The results of applying meta-dropout in Table.~\ref{table:4} are different from that in Table.~\ref{table:2} and Table.~\ref{table:3}. The reason is that meta-dropout is implemented by our best setting (dropblock) in Table.~\ref{table:4}. In table.~\ref{table:2} and Table.~\ref{table:3}, the normal dropout is used for showing the effectiveness of our idea in fair.

\subsubsection{The Influence of Meta-dropout on Base and Novel Classes}
Based on TFA, we study the influence of meta-dropout on the performance of base and novel classes. The results shown in Table.~\ref{table:100} prove that meta-dropout improves the accuracy of novel classes without hurting the performance of base classes.

\section{Few-shot Classification}

\subsection{Datasets}
We follow the commonly used datasets~\cite{chen2019closer,yang2021free}, such as Caltech-UCSD Birds-200-2011 (CUB) and mini-ImageNet in few-shot classification. CUB~\cite{wah2011multiclass} is for fine-grained classification, which contains 200 classes. We follow the evaluation protocol of~\cite{hilliard2018few} that 200 classes are divided into 100 base, 50 validation, and 50 novel classes, respectively. The mini-ImageNet~\cite{vinyals2016matching} that consists of 100 categories is a subset of ImageNet, and each class contains 600 images of size 84$\times$84. Follow the way of splitting the dataset in~\cite{finn2017model,ravi2016optimization,snell2017prototypical,sung2018learning}, the selected 100 classes are divided into 64 training classes, 16 validation classes, and 20 test classes. 


\begin{table*}[t]
	\setlength{\tabcolsep}{1.5pt}		
	\begin{center}
		\caption{Few-shot classification results for both the mini-ImageNet and CUB datasets. }
		\label{table:13}
		\begin{tabular}{cccccc}
			\hline\noalign{\smallskip}
			&&\multicolumn{2}{c}{CUB} & 	\multicolumn{2}{c}{mini-ImageNet}\\
			\hline\noalign{\smallskip}
			Method&backbone& 1-shot & 5-shot & 1-shot & 5-shot \\
			\noalign{\smallskip}
			\hline
			\noalign{\smallskip}
			MatchingNet~\cite{vinyals2016matching}&Conv-4&60.52$\pm$0.88&75.29$\pm$0.75 &48.14$\pm$0.78 &63.48$\pm$0.66\\
			ProtoNet~\cite{snell2017prototypical}&Conv-4&50.46$\pm$0.88& 76.39$\pm$0.64& 44.42$\pm$0.84 &64.24$\pm$0.72\\
			MAML~\cite{finn2017model}&Conv-4&54.73$\pm$0.97 &75.75$\pm$0.76 &46.47$\pm$0.82& 62.71$\pm$0.71\\
			RelationNet~\cite{sung2018learning}&Conv-4&62.34$\pm$0.94 &77.84$\pm$0.68& 49.31$\pm$0.85& 66.60$\pm$0.69\\
			Baseline~\cite{chen2019closer}&Conv-4&47.12$\pm$0.74 &64.16$\pm$0.71 &42.11$\pm$0.71& 62.53$\pm$0.69\\
			Baseline++~\cite{chen2019closer}&Conv-4&60.53$\pm$0.83 &79.34$\pm$0.61&48.24$\pm$0.75& 66.43$\pm$0.63\\
			Baseline++ (Our Impl.)&Conv-4&60.95$\pm$0.87&78.38$\pm$0.63&47.60$\pm$0.73&65.74$\pm$0.64\\
			Baseline++*&Conv-4&{\bf63.61$\pm$0.92}&{\bf80.00$\pm$0.62}&{\bf50.88$\pm$0.74}&{\bf68.27$\pm$0.65}\\
			\hline
			\noalign{\smallskip}
			Baseline++(Our Impl.) &ResNet-10&63.27$\pm$0.97&80.24$\pm$0.59&53.36$\pm$0.79&73.85$\pm$0.64\\
			Baseline++*&ResNet-10&{\bf69.05$\pm$0.88}&{\bf82.92$\pm$0.54}&{\bf56.34$\pm$0.78}&{\bf75.58$\pm$0.59}\\	
			\hline
		\end{tabular}
	\end{center}
\end{table*}

\subsection{Implementation Details}

We choose the Baseline++~\cite{chen2019closer}, which is a representative few-shot classification method, as our baseline. Based on this basic framework, we can prove the effectiveness of our meta-dropout in few-shot classification. During experiments, we follow the same training strategies in ~\cite{chen2019closer}. Specifically, Baseline++ is trained 200 epochs for the CUB dataset, and 400 epochs for the mini-ImageNet dataset. 
The evaluation setting we used is the same as the Baseline++. (600 randomly
episodes with the 95$\%$ conffdence intervals) 

\subsection{Comparison with Baselines}
We report the accuracy of few-shot classification in Table.~\ref{table:13}, in which we build our Baseline++* by applying meta-dropout on Baseline++. When testing, we use dropblock which is applied on the last convolution layer. 
The results demonstrate that our model outperforms other algorithms and using meta-dropout can significantly improve the performance of Baseline++ in almost all settings.

\subsection{Ablation Study}

\subsubsection{Meta-dropout vs Dropout}
We apply meta-dropout and dropout to show the importance of improving the generalization power of meta-knowledge. Using normal dropout to implement our meta-dropout for fair comparison.  The results are shown in Table.~\ref{table:15}. In the 5-shot setting of the CUB dataset, using meta-dropout is comparable to the Baseline++. However, applying meta-dropout can achieve the best performance in all other settings. 

\begin{table}[t]
	\setlength{\tabcolsep}{4pt}		
	\begin{center}
		\caption{Results of Baseline++ on the CUB and mini-ImageNet datasets by applying meta-dropout (M) and dropout (D). }
		\label{table:15}
		\begin{tabular}{ccccc}
			\hline\noalign{\smallskip}
			&\multicolumn{2}{c}{CUB} & 	\multicolumn{2}{c}{mini-ImageNet}\\
			\hline\noalign{\smallskip}
			Method& 1-shot & 5-shot & 1-shot & 5-shot \\
			\noalign{\smallskip}
			\hline
			\noalign{\smallskip}
			Baseline++ (Our Impl.)&60.95$\pm$0.87&{\bf78.38$\pm$0.63}&47.60$\pm$0.73&65.74$\pm$0.64\\
			+ M$\&$D&58.34$\pm$0.88&75.76$\pm$0.64&44.67$\pm$0.68&63.80$\pm$0.66\\
			+ D&59.05$\pm$0.87&76.39$\pm$0.65&43.92$\pm$0.69&62.66$\pm$0.67\\
			+ M&{\bf62.71$\pm$0.87}&78.12$\pm$0.62&{\bf50.47$\pm$0.72}&{\bf68.20$\pm$0.65}\\		
			\hline
		\end{tabular}
	\end{center}
\end{table}

\section{Conclusion}
In this paper, firstly,  we introduce a unified meta-learning framework, which explains two very different streams of few-shot learning, \emph{i.e.}, the episode-based and pre-train finetune-based few-shot learning. Secondly, we propose a simple, general, and effective meta-dropout to improve the generalization power of meta-knowledge in our framework. Finally, we conduct extensive experiments on the challenging few-shot object detection and few-shot image classification tasks, in which our model demonstrates great superiority towards the current excellent few-shot learning methods. We believe that our framework can provide more important insights and the proposed meta-dropout will be widely used in few-shot learning.

It is true that considering other techniques will make our paper look more complete. However, the goal of this paper is to show a new direction for few-shot models, and we can not include so much information in one paper. The new research direction is exploring the power of regularization techniques on few-shot models.

\nocite{langley00}

\bibliographystyle{splncs}
\bibliography{egbib}

\begin{thebibliography}{10}

\bibitem{ren2016faster}
Ren, S., He, K., Girshick, R., Sun, J.:
\newblock Faster r-cnn: Towards real-time object detection with region proposal
  networks.
\newblock IEEE transactions on pattern analysis and machine intelligence
  \textbf{39} (2016)  1137--1149

\bibitem{dai2016r}
Dai, J., Li, Y., He, K., Sun, J.:
\newblock R-fcn: Object detection via region-based fully convolutional
  networks.
\newblock arXiv preprint arXiv:1605.06409 (2016)

\bibitem{redmon2017yolo9000}
Redmon, J., Farhadi, A.:
\newblock Yolo9000: better, faster, stronger.
\newblock In: Proceedings of the IEEE conference on computer vision and pattern
  recognition. (2017)  7263--7271

\bibitem{lin2017feature}
Lin, T.Y., Doll{\'a}r, P., Girshick, R., He, K., Hariharan, B., Belongie, S.:
\newblock Feature pyramid networks for object detection.
\newblock In: Proceedings of the IEEE conference on computer vision and pattern
  recognition. (2017)  2117--2125

\bibitem{kang2019few}
Kang, B., Liu, Z., Wang, X., Yu, F., Feng, J., Darrell, T.:
\newblock Few-shot object detection via feature reweighting.
\newblock In: Proceedings of the IEEE International Conference on Computer
  Vision. (2019)  8420--8429

\bibitem{yan2019meta}
Yan, X., Chen, Z., Xu, A., Wang, X., Liang, X., Lin, L.:
\newblock Meta r-cnn: Towards general solver for instance-level low-shot
  learning.
\newblock In: Proceedings of the IEEE International Conference on Computer
  Vision. (2019)  9577--9586

\bibitem{wang2019meta}
Wang, Y.X., Ramanan, D., Hebert, M.:
\newblock Meta-learning to detect rare objects.
\newblock In: Proceedings of the IEEE International Conference on Computer
  Vision. (2019)  9925--9934

\bibitem{fan2020few}
Fan, Q., Zhuo, W., Tang, C.K., Tai, Y.W.:
\newblock Few-shot object detection with attention-rpn and multi-relation
  detector.
\newblock In: Proceedings of the IEEE/CVF Conference on Computer Vision and
  Pattern Recognition. (2020)  4013--4022

\bibitem{dhillon2019baseline}
Dhillon, G.S., Chaudhari, P., Ravichandran, A., Soatto, S.:
\newblock A baseline for few-shot image classification.
\newblock arXiv preprint arXiv:1909.02729 (2019)

\bibitem{chen2019closer}
Chen, W.Y., Liu, Y.C., Kira, Z., Wang, Y.C.F., Huang, J.B.:
\newblock A closer look at few-shot classification.
\newblock arXiv preprint arXiv:1904.04232 (2019)

\bibitem{wang2020frustratingly}
Wang, X., Huang, T.E., Darrell, T., Gonzalez, J.E., Yu, F.:
\newblock Frustratingly simple few-shot object detection.
\newblock arXiv preprint arXiv:2003.06957 (2020)

\bibitem{vinyals2016matching}
Vinyals, O., Blundell, C., Lillicrap, T., Wierstra, D.,  et~al.:
\newblock Matching networks for one shot learning.
\newblock Advances in neural information processing systems \textbf{29} (2016)
  3630--3638

\bibitem{sun2021fsce}
Sun, B., Li, B., Cai, S., Yuan, Y., Zhang, C.:
\newblock Fsce: Few-shot object detection via contrastive proposal encoding.
\newblock In: Proceedings of the IEEE/CVF Conference on Computer Vision and
  Pattern Recognition. (2021)  7352--7362

\bibitem{hariharan2017low}
Hariharan, B., Girshick, R.:
\newblock Low-shot visual recognition by shrinking and hallucinating features.
\newblock In: Proceedings of the IEEE International Conference on Computer
  Vision. (2017)  3018--3027

\bibitem{koch2015siamese}
Koch, G., Zemel, R., Salakhutdinov, R.:
\newblock Siamese neural networks for one-shot image recognition.
\newblock In: ICML deep learning workshop. Volume~2., Lille (2015)

\bibitem{tokmakov2019learning}
Tokmakov, P., Wang, Y.X., Hebert, M.:
\newblock Learning compositional representations for few-shot recognition.
\newblock In: Proceedings of the IEEE International Conference on Computer
  Vision. (2019)  6372--6381

\bibitem{andrychowicz2016learning}
Andrychowicz, M., Denil, M., Gomez, S., Hoffman, M.W., Pfau, D., Schaul, T.,
  Shillingford, B., De~Freitas, N.:
\newblock Learning to learn by gradient descent by gradient descent.
\newblock In: Advances in neural information processing systems. (2016)
  3981--3989

\bibitem{munkhdalai2017meta}
Munkhdalai, T., Yu, H.:
\newblock Meta networks.
\newblock Proceedings of machine learning research \textbf{70} (2017)  2554

\bibitem{santoro2016one}
Santoro, A., Bartunov, S., Botvinick, M., Wierstra, D., Lillicrap, T.:
\newblock One-shot learning with memory-augmented neural networks.
\newblock arXiv preprint arXiv:1605.06065 (2016)

\bibitem{thrun1998lifelong}
Thrun, S.:
\newblock Lifelong learning algorithms.
\newblock In: Learning to learn.
\newblock Springer (1998)  181--209

\bibitem{snell2017prototypical}
Snell, J., Swersky, K., Zemel, R.:
\newblock Prototypical networks for few-shot learning.
\newblock In: Advances in neural information processing systems. (2017)
  4077--4087

\bibitem{gidaris2018dynamic}
Gidaris, S., Komodakis, N.:
\newblock Dynamic few-shot visual learning without forgetting.
\newblock In: Proceedings of the IEEE Conference on Computer Vision and Pattern
  Recognition. (2018)  4367--4375

\bibitem{sung2018learning}
Sung, F., Yang, Y., Zhang, L., Xiang, T., Torr, P.H., Hospedales, T.M.:
\newblock Learning to compare: Relation network for few-shot learning.
\newblock In: Proceedings of the IEEE Conference on Computer Vision and Pattern
  Recognition. (2018)  1199--1208

\bibitem{kim2019edge}
Kim, J., Kim, T., Kim, S., Yoo, C.D.:
\newblock Edge-labeling graph neural network for few-shot learning.
\newblock In: Proceedings of the IEEE Conference on Computer Vision and Pattern
  Recognition. (2019)  11--20

\bibitem{gidaris2019generating}
Gidaris, S., Komodakis, N.:
\newblock Generating classification weights with gnn denoising autoencoders for
  few-shot learning.
\newblock In: Proceedings of the IEEE Conference on Computer Vision and Pattern
  Recognition. (2019)  21--30

\bibitem{karlinsky2019repmet}
Karlinsky, L., Shtok, J., Harary, S., Schwartz, E., Aides, A., Feris, R.,
  Giryes, R., Bronstein, A.M.:
\newblock Repmet: Representative-based metric learning for classification and
  few-shot object detection.
\newblock In: Proceedings of the IEEE Conference on Computer Vision and Pattern
  Recognition. (2019)  5197--5206

\bibitem{chen2021meta}
Chen, Y., Liu, Z., Xu, H., Darrell, T., Wang, X.:
\newblock Meta-baseline: Exploring simple meta-learning for few-shot learning.
\newblock In: Proceedings of the IEEE/CVF International Conference on Computer
  Vision. (2021)  9062--9071

\bibitem{liu2016ssd}
Liu, W., Anguelov, D., Erhan, D., Szegedy, C., Reed, S., Fu, C.Y., Berg, A.C.:
\newblock Ssd: Single shot multibox detector.
\newblock In: European conference on computer vision, Springer (2016)  21--37

\bibitem{srivastava2014dropout}
Srivastava, N., Hinton, G., Krizhevsky, A., Sutskever, I., Salakhutdinov, R.:
\newblock Dropout: a simple way to prevent neural networks from overfitting.
\newblock The journal of machine learning research \textbf{15} (2014)
  1929--1958

\bibitem{triantafillou2019meta}
Triantafillou, E., Zhu, T., Dumoulin, V., Lamblin, P., Evci, U., Xu, K.,
  Goroshin, R., Gelada, C., Swersky, K., Manzagol, P.A.,  et~al.:
\newblock Meta-dataset: A dataset of datasets for learning to learn from few
  examples.
\newblock arXiv preprint arXiv:1903.03096 (2019)

\bibitem{everingham2015pascal}
Everingham, M., Eslami, S.A., Van~Gool, L., Williams, C.K., Winn, J.,
  Zisserman, A.:
\newblock The pascal visual object classes challenge: A retrospective.
\newblock International journal of computer vision \textbf{111} (2015)  98--136

\bibitem{everingham2010pascal}
Everingham, M., Van~Gool, L., Williams, C.K., Winn, J., Zisserman, A.:
\newblock The pascal visual object classes (voc) challenge.
\newblock International journal of computer vision \textbf{88} (2010)  303--338

\bibitem{lin2014microsoft}
Lin, T.Y., Maire, M., Belongie, S., Hays, J., Perona, P., Ramanan, D.,
  Doll{\'a}r, P., Zitnick, C.L.:
\newblock Microsoft coco: Common objects in context.
\newblock In: European conference on computer vision, Springer (2014)  740--755

\bibitem{li2021beyond}
Li, B., Yang, B., Liu, C., Liu, F., Ji, R., Ye, Q.:
\newblock Beyond max-margin: Class margin equilibrium for few-shot object
  detection.
\newblock In: Proceedings of the IEEE/CVF Conference on Computer Vision and
  Pattern Recognition. (2021)  7363--7372

\bibitem{wu2020multi}
Wu, J., Liu, S., Huang, D., Wang, Y.:
\newblock Multi-scale positive sample refinement for few-shot object detection.
\newblock In: European Conference on Computer Vision, Springer (2020)  456--472

\bibitem{li2021few}
Li, Y., Zhu, H., Cheng, Y., Wang, W., Teo, C.S., Xiang, C., Vadakkepat, P.,
  Lee, T.H.:
\newblock Few-shot object detection via classification refinement and
  distractor retreatment.
\newblock In: Proceedings of the IEEE/CVF Conference on Computer Vision and
  Pattern Recognition. (2021)  15395--15403

\bibitem{fan2021generalized}
Fan, Z., Ma, Y., Li, Z., Sun, J.:
\newblock Generalized few-shot object detection without forgetting.
\newblock In: Proceedings of the IEEE/CVF Conference on Computer Vision and
  Pattern Recognition. (2021)  4527--4536

\bibitem{zhang2021hallucination}
Zhang, W., Wang, Y.X.:
\newblock Hallucination improves few-shot object detection.
\newblock In: Proceedings of the IEEE/CVF Conference on Computer Vision and
  Pattern Recognition. (2021)  13008--13017

\bibitem{ghiasi2018dropblock}
Ghiasi, G., Lin, T.Y., Le, Q.V.:
\newblock Dropblock: A regularization method for convolutional networks.
\newblock arXiv preprint arXiv:1810.12890 (2018)

\bibitem{tompson2015efficient}
Tompson, J., Goroshin, R., Jain, A., LeCun, Y., Bregler, C.:
\newblock Efficient object localization using convolutional networks.
\newblock In: Proceedings of the IEEE conference on computer vision and pattern
  recognition. (2015)  648--656

\bibitem{yang2021free}
Yang, S., Liu, L., Xu, M.:
\newblock Free lunch for few-shot learning: Distribution calibration.
\newblock arXiv preprint arXiv:2101.06395 (2021)

\bibitem{wah2011multiclass}
Wah, C., Branson, S., Perona, P., Belongie, S.:
\newblock Multiclass recognition and part localization with humans in the loop.
\newblock In: 2011 International Conference on Computer Vision, IEEE (2011)
  2524--2531

\bibitem{hilliard2018few}
Hilliard, N., Phillips, L., Howland, S., Yankov, A., Corley, C.D., Hodas, N.O.:
\newblock Few-shot learning with metric-agnostic conditional embeddings.
\newblock arXiv preprint arXiv:1802.04376 (2018)

\bibitem{finn2017model}
Finn, C., Abbeel, P., Levine, S.:
\newblock Model-agnostic meta-learning for fast adaptation of deep networks.
\newblock arXiv preprint arXiv:1703.03400 (2017)

\bibitem{ravi2016optimization}
Ravi, S., Larochelle, H.:
\newblock Optimization as a model for few-shot learning.
\newblock (2016)

\end{thebibliography}


\newpage
\appendix
\onecolumn
\section{Appendix}

\subsection{Few-shot Object Detection}

\begin{table*}[t]
	\setlength{\tabcolsep}{4pt}		
	\begin{center}
		\caption{Results of mAP of novel classes on the split 1 of VOC 2007 test set by adopting different types of meta-dropout on Meta R-CNN and TFA.}
		\label{table:8}
	    \begin{tabular}{ccccccc}
		\hline\noalign{\smallskip}
		&\multicolumn{3}{c}{Meta R-CNN}&\multicolumn{3}{c}{TFA}\\
		\hline\noalign{\smallskip}
		Setting / Shot & 1 & 3 & 10&1&3&10\\
		\noalign{\smallskip}
		\hline
		\noalign{\smallskip}
		Baseline&14.7&32.8&{\bf51.9}&22.4&40.3&53.1\\
		+ dropout &14.3&31.3&50.2&25.1&41.9&53\\
		+ dropblock &{\bf16.2}&{\bf32.9}&51.4&{\bf25.9}&{\bf42.5}&{\bf54.5}\\
		\hline
		\end{tabular}
	\end{center}
\end{table*}

\subsubsection{Implementation of Meta-dropout}
We study the influence of the implementation of meta-dropout on performance. For example, our meta-dropout can be implemented by using normal dropout or dropblock. When using the Meta R-CNN framework, we apply meta-dropout on group 4 of the backbone, while meta-dropout is applied on the last convolution layer of TFA. The experimental results are shown in Table.~\ref{table:8}. In general, using dropblock to implement our meta-dropout can get higher accuracy in novel classes.

\subsubsection{Specific Locations of Applying Meta-dropout}

\begin{table*}[t]
	\setlength{\tabcolsep}{4pt}		
	\begin{center}
		\caption{Results of novel mAP on the split 1 of VOC 2007 test set by adopting meta-dropout on different locations of Meta R-CNN.}
		\label{table:6}
		\begin{tabular}{cccc}
			\hline\noalign{\smallskip}
			Setting / Shot (bs=1/4)& 1 & 3 & 10\\
			\noalign{\smallskip}
			\hline
			\noalign{\smallskip}
			Meta R-CNN&23.3/14.7&37.8/32.8&49.3/{\bf51.9}\\
			on last conv layer &20.3/15.6&31.6/32.2&46.3/50.8\\
			on group 4 &24.5/{\bf16.2}&36/{\bf32.9}&50/51.4\\
			on group 3$\&$4 &{\bf24.7}/16.1&{\bf37.3}/32.5&{\bf51.8}/50\\
			\hline
		\end{tabular}
	\end{center}
\end{table*}

\begin{table*}[t]
	\setlength{\tabcolsep}{4pt}		
	\begin{center}
		\caption{Results of novel mAP on the VOC 2007 test set by adopting meta-dropout on different locations of TFA.}
		\label{table:7}
		\begin{tabular}{ccccccc}
			\hline\noalign{\smallskip}
			&\multicolumn{3}{c}{split 1} & 	\multicolumn{3}{c}{split 2}\\
			\hline\noalign{\smallskip}
			Setting / Shot& 1 & 3 & 10 & 1 & 3 & 10\\
			\noalign{\smallskip}
			\hline
			\noalign{\smallskip}
			TFA&22.4&40.3&53.1&15.6&26.7&37.4\\
			on last conv layer &25.9&42.5&54.5&{\bf17.3}&28.6&39.9\\
			on group 4 &{\bf26.7}&44&54.2&14.3&28.3&39.3\\
			on group 3$\&$4 &26.3&{\bf45.6}&{\bf55.8}&15.9&{\bf29.9}&{\bf40.8}\\
			\hline
		\end{tabular}
	\end{center}
\end{table*}

Selecting Meta R-CNN and TFA as our baselines. In particular, dropblock is used to implement meta-dropout. From Table.~\ref{table:6}, we conclude that applying meta-dropout on group 4 of the backbone is better than on group 3$\&$4 with the setting of batch size 4. While using meta-dropout on group 3$\&$4 achieves higher performance when batch size is 1. The results of applying meta-dropout on TFA are shown in Table.~\ref{table:7}, in which using meta-dropout on group 3$\&$4 gets the highest performance in most settings.
The keep prob and block size are two important hyper-parameters in dropblock, which are set to 0.9 and 7 separately. Specifically, when using meta-dropout on group 3 or 4 of the backbone network, meta-dropout is applied to the last convolution layer of each bottleneck block.

\subsubsection{The influence of Meta-dropout on Base and Novel classes}
Based on the original setting of TFA, we study the influence of meta-dropout on the performance of base and novel classes. The results shown in Table.~\ref{table:100} prove that our method can improve the accuracy of novel classes by a large margin without hurting the performance of base classes.

\begin{table}[t]
	\setlength{\tabcolsep}{3pt}	
	\begin{center}
		\caption{Results of base and novel mAP on the split 1\&2 of VOC 2007 test set by adopting meta-dropout on TFA. }
		\label{table:100}
		\begin{tabular}{ccccccc}
			\hline\noalign{\smallskip}
			&\multicolumn{3}{c}{split 1} & 	\multicolumn{3}{c}{split2}\\
			\hline\noalign{\smallskip}
			Base / Novel & 1 & 3 &10 & 1 & 3&10 \\
			\hline
			\noalign{\smallskip}
			TFA&79.8/37.2&79.1/{\bf44.4}&{\bf79.4}/53.4&79.8/{\bf21.6}&78.6/34.8&78.3/38.4\\
			TFA*&{\bf80}/{\bf39.7}&{\bf79.7}/43.4&79.2/ {\bf55.2}&{\bf80.6}/20.3&{\bf78.8}/{\bf36.6}&{\bf78.7}/{\bf41.8}\\					
			\hline
		\end{tabular}
	\end{center}
\end{table}

\subsubsection{Batch Size}

We explore the influence of batch size on Meta R-CNN and TFA. The experimental results are shown in Table.~\ref{table:10}. 
Generally, on Meta R-CNN, as the batch size gets smaller, the performance is higher. Contrary to Meta R-CNN, as the batch size gets larger, the performance is higher in TFA, which is consistent with the previous experimental intuition.

\begin{table}[t]
	\begin{center}
		\caption{Results of overall and novel mAP on the split 1 of VOC 2007 test set by adopting different batch size (bs) on Meta R-CNN and TFA. }
		\label{table:10}
		\begin{tabular}{cccccc}
			\hline\noalign{\smallskip}
			&&\multicolumn{2}{c}{Overall mAP} & 	\multicolumn{2}{c}{Novel mAP}\\
			\hline\noalign{\smallskip}
			Methods&Bs / Shot& 1 & 3  & 1 & 3 \\
			\hline
			\noalign{\smallskip}
			\multirow{3}*{Meta R-CNN}&1&{\bf53.17}&{\bf59.46}&{\bf23.3}&{\bf37.83}\\
			&2&50.1&58.09&20.81&35.43\\
			&4&43.9&55.73&14.74&32.83\\
			\hline
			\noalign{\smallskip}
			\multirow{3}*{TFA}&4&55.4&58.2&17.2&30.8\\
			&8&62.2&66.3&{\bf23.5}&39.8\\
			&16&{\bf63.5}&{\bf67.8}&22.4&{\bf40.3}\\					
			\hline
		\end{tabular}
	\end{center}
\end{table}

\subsection{Few-shot Classification}

\subsubsection{Implementation of Meta-dropout}
We study the way of implementing our meta-dropout by using dropblock with different block sizes and normal dropout with different locations. Due to the flatten layer is 1-dimensional, dropblock can only be used on the last convolution layer. Normal dropout can be applied on the last convolution or the last flatten layer. In Table.~\ref{table:16}, the experimental results show that using dropblock with block size 7 is suitable for the CUB dataset, and applying normal dropout on the flatten layer achieves the best performance on the mini-ImageNet dataset.

\begin{table}[t]
	\setlength{\tabcolsep}{4pt}		
	\begin{center}
		\caption{Results on the CUB and mini-ImageNet datasets by applying different types of meta-dropout on Baseline++. $bl$ is the block size in dropblock.}
		\label{table:16}
		\begin{tabular}{ccccc}
			\hline\noalign{\smallskip}
			&\multicolumn{2}{c}{CUB} & 	\multicolumn{2}{c}{mini-ImageNet}\\
			\hline\noalign{\smallskip}
			Method& 1-shot & 5-shot & 1-shot & 5-shot \\
			\noalign{\smallskip}
			\hline
			\noalign{\smallskip}
			Baseline++ (Our Impl.)&60.95$\pm$0.87&78.38$\pm$0.63&47.60$\pm$0.73&65.74$\pm$0.64\\
			dropblock ($bl$=3)&62.03$\pm$0.86&79.70$\pm$0.60&48.22$\pm$0.77&67.06$\pm$0.66\\
			dropblock ($bl$=7)&{\bf63.61$\pm$0.92}&{\bf80.00$\pm$0.62}&47.60$\pm$0.71&67.25$\pm$0.62\\
			dropout$\_$conv&62.45$\pm$0.91&78.70$\pm$0.63&48.83$\pm$0.75&67.11$\pm$0.63\\
			dropout$\_$flatten&62.71$\pm$0.87&78.12$\pm$0.62&{\bf50.47$\pm$0.72}&{\bf68.20$\pm$0.65}\\		
			\hline
		\end{tabular}
	\end{center}
\end{table}

\subsubsection{Specific Locations of Applying Meta-dropout}
For exploring the effect of the specific location of applying meta-dropout in the backbone, we use normal dropout as our meta-dropout, and apply it on the last convolution layer and the last flatten layer. The results in Table.~\ref{table:17} show that applying meta-dropout on the last 1-dimensional feature is better than on the last convolution layer.

\begin{table}[t]
	\setlength{\tabcolsep}{4pt}		
	\begin{center}
		\caption{Results on the CUB and mini-ImageNet datasets by applying meta-dropout on different locations of Baseline++. }
		\label{table:17}
		\begin{tabular}{ccccc}
			\hline\noalign{\smallskip}
			&\multicolumn{2}{c}{CUB} & 	\multicolumn{2}{c}{mini-ImageNet}\\
			\hline\noalign{\smallskip}
			Method& 1-shot & 5-shot & 1-shot & 5-shot \\
			\noalign{\smallskip}
			\hline
			\noalign{\smallskip}
			Baseline++ (Our Impl.)&60.95$\pm$0.87&78.38$\pm$0.63&47.60$\pm$0.73&65.74$\pm$0.64\\
			on last conv layer&62.45$\pm$0.91&{\bf78.70$\pm$0.63}&48.83$\pm$0.75&67.11$\pm$0.63\\
			on last flatten layer&{\bf62.71$\pm$0.87}&78.12$\pm$0.62&{\bf50.47$\pm$0.72}&{\bf68.20$\pm$0.65}\\		
			\hline
		\end{tabular}
	\end{center}
\end{table}

\subsubsection{Batch Size}

Based on the Baseline++~\cite{chen2019closer} with batch size 16, we conduct experiments to explore the influence of batch size, whose results are shown in Table.~\ref{table:18}. We find using 32 as the batch size is the best choice for Baseline++.

\begin{table}[b]
	\begin{center}
		\caption{Results of Baseline++ on the CUB and mini-ImageNet datasets by adopting different batch size. }
		\label{table:18}
		\begin{tabular}{ccccc}
			\hline\noalign{\smallskip}
			&\multicolumn{2}{c}{CUB} & 	\multicolumn{2}{c}{mini-ImageNet}\\
			\hline\noalign{\smallskip}
			Batch size& 1-shot & 5-shot & 1-shot & 5-shot \\
			\noalign{\smallskip}
			\hline
			\noalign{\smallskip}
			16&60.95$\pm$0.87&78.38$\pm$0.63&47.60$\pm$0.73&65.74$\pm$0.64\\
			32&63.63$\pm$0.88&{\bf78.96$\pm$0.63}&{\bf50.21$\pm$0.75}&{\bf68.23$\pm$0.67}\\
			64&{\bf63.67$\pm$0.88}&78.78$\pm$0.65&48.77$\pm$0.71&67.82$\pm$0.66\\
			128&63.13$\pm$0.93&78.27$\pm$0.63&49.59$\pm$0.74&68.04$\pm$0.66\\
			\hline
		\end{tabular}
	\end{center}
\end{table}



\end{document}